\documentclass[11pt,a4paper]{article}

\usepackage[utf8]{inputenc}
\usepackage[T1]{fontenc}
\usepackage{amsmath,amsfonts,amssymb}
\usepackage{graphicx}
\usepackage{booktabs}
\usepackage{cite}
\usepackage{geometry}
\usepackage{lmodern}
\usepackage{microtype}
\usepackage[hyphens]{url}
\usepackage{hyperref}
\usepackage{xcolor}
\usepackage{setspace}

\hypersetup{
  colorlinks=true,
  linkcolor=blue!80!black,
  citecolor=blue!80!black,
  urlcolor=blue!80!black
}

\title{\textbf{\Large Breakdown of Autoregressive Multi-Step Reasoning Under Concentrated vs.\ Diffuse Attention Head Dropout}}

\author{
  \textbf{Karol Struniawski}\thanks{Email: \texttt{karol\_struniawski@sggw.edu.pl}} \\
  \small Institute of Information Technology, Warsaw University of Life Sciences -- SGGW \\
  \small Warsaw, Poland
}

\date{}

\begin{document}

\maketitle
\doublespacing

\begin{abstract}
During extended wakefulness, mammalian neocortical networks experience localized periods of transient neuronal silence termed off-states. Here, we examine how analogous spatiotemporally concentrated disruptions affect sequential reasoning in autoregressive Large Language Models (LLMs). Evaluating 250 paired derivations on the GSM8K benchmark across two 8B models (Qwen3-8B and LLaMA-3.1-8B-Instruct), we contrast clustered bursts of attention head masking against budget-matched diffuse i.i.d. dropout. Concentrated bursts induce substantial reasoning degradation: at onset parameter $\mathcal{B}=0.10$ (scheduling $\sim 5.7\%$ of vulnerable head-token volume), Qwen3-8B accuracy drops from $91.2\%$ under diffuse noise to $78.8\%$ under burst dropout ($\Delta = -12.4$ percentage points [pp], exact McNemar $p = 5.5 \times 10^{-6}$), while LLaMA-3.1-8B-Instruct drops from $84.0\%$ to $76.8\%$ ($\Delta = -7.2$ pp, $p = 0.0198$). Conversely, dispersing the identical cumulative budget uniformly across generation steps via stratified quintile matching yields substantially milder degradation (retaining $91.2\%$ in Qwen and $84.0\%$ in LLaMA). In an exploratory factorial dissociation ($N=100$), single-step dropout density is associated with an acute drop in a single schedule draw ($-10.0$ pp on Qwen, $p_{\text{holm}} = 0.0425$), while multi-token continuity exhibits an inconclusive compounding trend ($-7.0$ pp, $p_{\text{holm}} = 0.1892$). Error analysis reveals that Qwen failures reflect derivation errors within formatted responses (conditional accuracy drop of $-11.4$ pp), whereas in LLaMA $22.2\%$ of discordant losses lack a final answer format line (conditional gap narrows to $-5.0$ pp). These findings demonstrate that autoregressive reasoning chains are particularly vulnerable to concentrated structural disruptions, providing empirical benchmarks for fault tolerance in transformer architectures.
\end{abstract}

\newpage

\section*{Introduction}

In biological neuroscience, the classical view of sleep as an exclusively global brain state mediated solely by subcortical arousal nuclei has been revised following the discovery of local sleep in awake neocortical networks \cite{vyazovskiy2011local, nir2017selective, tononi2014sleep}. Under extended wakefulness and sleep deprivation, individual cortical columns and localized neuronal assemblies spontaneously undergo transient periods of neuronal silence and slow-wave electrophysiological activity, termed cortical off-states or microsleeps, even as the organism remains awake and active \cite{vyazovskiy2011local, krueger2008sleep}. In awake humans, these intruding localized off-states predict cognitive lapses during tasks that require continuous, sequential mental operations \cite{nir2017selective, andrillon2021predicting}.

In artificial intelligence, Large Language Models (LLMs) parameterized by deep transformer architectures have proven effective at solving multi-step logical tasks \cite{vaswani2017attention, wei2022chain, kojima2022large}. In Chain-of-Thought (CoT) reasoning, complex problems are decomposed into sequential token trajectories. Intermediate arithmetic calculations, variable groundings, and inferential deductions are generated autoregressively and appended to the context history \cite{cobbe2021gsm8k, dziri2023faith, wang2023selfconsistency}. Attention heads act as dynamic routing channels, reading relevant historical representations from the Key--Value (KV) cache to compute successive tokens \cite{olsson2022context, elhage2021mathematical}.

While the fault tolerance of transformer attention has been analyzed in the context of post-training structural pruning \cite{michel2019sixteen, voita2019analyzing}, isotropic weight degradation, and uniform pre-training dropout \cite{srivastava2014dropout}, dynamic runtime noise models commonly adopt an independent and identically distributed (i.i.d.) perturbation schedule. Under uniform i.i.d.\ perturbations, omissions are dispersed uniformly at random across the spatio-temporal volume of the network. However, autoregressive computation exhibits an unrolled temporal dependency structure, where the output token generated at step $t$ conditions all subsequent representations at steps $t+1, t+2, \dots, T$.

This structural asymmetry raises two practical questions regarding the robustness of autoregressive reasoning: first, whether spatiotemporally concentrated attention head failure (co-activating head omissions across layers and successive tokens) inflicts a more severe reasoning breakdown than an equal cumulative budget of diffuse (i.i.d.) noise, and second, whether the resulting impairment is driven primarily by instantaneous per-step intensity (dense head loss within a single token) or by temporal multi-token continuity (sustained disruption over consecutive tokens).

To evaluate these questions with paired experimental control, we conduct confirmatory evaluations ($N=250$ items) and an exploratory factorial dissociation experiment ($N=100$) on two modern 8-billion-parameter open-weight models: Qwen3-8B \cite{qwen2025qwen3} and LLaMA-3.1-8B-Instruct \cite{dubey2024llama} on the GSM8K benchmark \cite{cobbe2021gsm8k}. We show that concentrated bursts trigger significant derivation impairment ($\Delta = -12.4$ pp on Qwen, $\Delta = -7.2$ pp on LLaMA at nominal onset parameter $\mathcal{B}=0.10$), while matched budgets of diffuse i.i.d.\ noise result in comparatively modest losses. Furthermore, we evaluate instantaneous per-step density and multi-token continuity, finding that single-step concentration produces an acute drop in that draw ($-10.0$ pp on Qwen, $p_{\text{holm}}=0.0425$), with multi-token continuity acting as a suggestive compounding trend.

\section*{Results}

\subsection*{Confirmatory Evaluation: The Clustered Deficit}

\begin{figure}[htbp]
\centering
\includegraphics[width=0.98\textwidth]{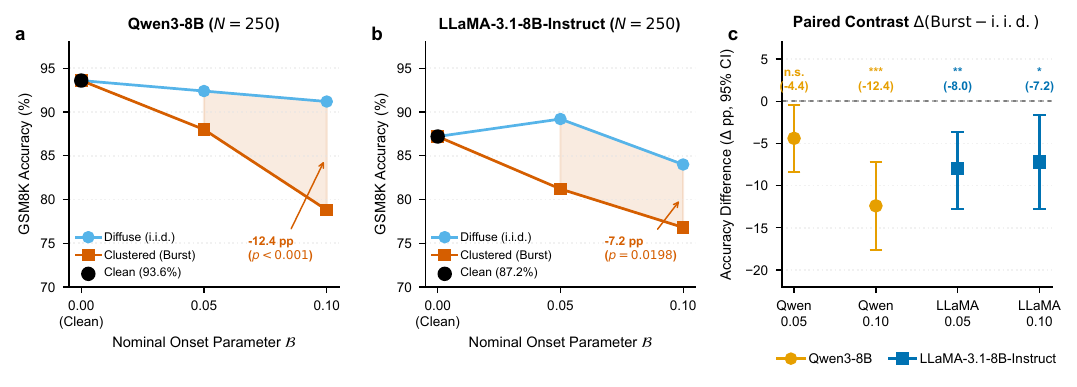}
\caption{\textbf{Derivation breakdown under concentrated attention head dropout contrasted with diffuse noise resilience ($N=250$ paired GSM8K derivations).} \textbf{a}, Qwen3-8B derivation accuracy under clean baseline ($93.6\%$, black), diffuse i.i.d.\ noise (sky blue), and clustered burst dropout (vermilion) across nominal onset parameters $\mathcal{B}=0.05$ and $\mathcal{B}=0.10$. At $\mathcal{B}=0.10$, burst dropout drops accuracy to $78.8\%$, whereas diffuse noise remains at $91.2\%$. \textbf{b}, LLaMA-3.1-8B-Instruct derivation accuracy under identical experimental conditions (Clean $87.2\% \to$ i.i.d.\ $84.0\%$ vs.\ Burst $76.8\%$). \textbf{c}, Paired difference contrast $\Delta(\text{Burst} - \text{i.i.d.})$ in percentage points (pp) with 95\% bootstrap confidence intervals. For Qwen3-8B at $\mathcal{B}=0.05$, the bootstrap CI excludes zero ($[-8.4\text{ pp}, -0.4\text{ pp}]$) whereas the exact McNemar test yields $p_{\text{holm}} = 0.0614$ (discordant pairs: 20 vs.\ 9). Asterisks denote family-wise Holm-Bonferroni significance: $^{***}\, p_{\text{holm}} < 0.001$, $^{**}\, p_{\text{holm}} < 0.01$, $^{*}\, p_{\text{holm}} < 0.05$, n.s.: not significant.}
\label{fig:headline}
\end{figure}

In the confirmatory evaluation across $N=250$ paired GSM8K derivations, we evaluated Qwen3-8B and LLaMA-3.1-8B-Instruct across onset parameters $\mathcal{B} \in \{0.05, 0.10\}$ (Fig.~\ref{fig:headline}, Table~\ref{tab:gate2}).

\begin{table}[htbp]
\centering
\footnotesize
\caption{\textbf{Confirmatory Results across $N=250$ paired derivations on GSM8K.} Performance metrics, realized head-token budgets, answer completion rates, raw task accuracy, accuracy conditional on emitting a final answer line, discordant pairs $(b, c)$ where $b=\text{i.i.d.}^+ \text{Burst}^-$ and $c=\text{Burst}^+ \text{i.i.d.}^-$, exact McNemar $p$-values, Holm-corrected $p_{\text{holm}}$, and 95\% bootstrap confidence intervals.}
\label{tab:gate2}
\vspace{2mm}
\resizebox{\textwidth}{!}{%
\begin{tabular}{llcccccccccc}
\toprule
Model & Arm & Budget $\mathcal{B}$ & Realized Heads & Answer Rate & Raw Acc & Cond Acc & Paired $\Delta$ & Discordant $(b, c)$ & $p_{\text{McNemar}}$ & $p_{\text{holm}}$ & 95\% Bootstrap CI \\
\midrule
Qwen3-8B & Clean & 0.00 & 0.0 & 99.6\% & 93.6\% & 94.0\% & Baseline & --- & --- & --- & --- \\
 & i.i.d. & 0.05 & 4,969.8 & 98.4\% & 92.4\% & 93.5\% & --- & --- & --- & --- & --- \\
 & Burst & 0.05 & 4,868.6 & 98.8\% & 88.0\% & 88.7\% & $-4.4$ pp & $(20, 9)$ & 0.0614 & 0.0614 & $[-8.4\text{ pp}, -0.4\text{ pp}]$ \\
 & i.i.d. & 0.10 & 8,631.3 & 98.8\% & 91.2\% & 91.9\% & --- & --- & --- & --- & --- \\
 & Burst & 0.10 & 8,897.3 & 96.4\% & 78.8\% & 80.5\% & $-12.4$ pp & $(39, 8)$ & $5.5 \times 10^{-6}$ & $1.1 \times 10^{-5}$ & $[-17.6\text{ pp}, -7.2\text{ pp}]$\\
\midrule
LLaMA-3.1-8B-Instruct & Clean & 0.00 & 0.0 & 98.8\% & 87.2\% & 87.9\% & Baseline & --- & --- & --- & --- \\
 & i.i.d. & 0.05 & 3,826.2 & 96.0\% & 89.2\% & 90.0\% & --- & --- & --- & --- & --- \\
 & Burst & 0.05 & 3,986.0 & 92.4\% & 81.2\% & 84.0\% & $-8.0$ pp & $(29, 9)$ & 0.0017 & 0.0033 & $[-12.8\text{ pp}, -3.6\text{ pp}]$ \\
 & i.i.d. & 0.10 & 6,437.2 & 94.0\% & 84.0\% & 85.1\% & --- & --- & --- & --- & --- \\
 & Burst & 0.10 & 6,626.4 & 90.4\% & 76.8\% & 80.1\% & $-7.2$ pp & $(36, 18)$ & 0.0198 & 0.0198 & $[-12.8\text{ pp}, -1.6\text{ pp}]$ \\
\bottomrule
\end{tabular}%
}
\end{table}

\paragraph{Qwen3-8B Derivations}
Under unperturbed clean inference, Qwen3-8B achieves an accuracy of $93.6\%$ ($234/250$). Diffuse i.i.d.\ dropout introduces minor disruption, with accuracy dropping slightly to $92.4\%$ at $\mathcal{B}=0.05$ ($-1.2$ pp) and $91.2\%$ at $\mathcal{B}=0.10$ ($-2.4$ pp).

Under concentrated burst dropout, Qwen3-8B exhibits clear vulnerability. At $\mathcal{B}=0.05$, accuracy declines to $88.0\%$ ($\Delta = -4.4$ pp, exact McNemar $p = 0.0614$, $p_{\text{holm}} = 0.0614$, 95\% CI: $[-8.4\text{ pp}, -0.4\text{ pp}]$). Here, the unadjusted bootstrap CI excludes zero while the exact McNemar test yields $p = 0.0614$. This divergence occurs because the percentile bootstrap resamples paired item differences with replacement, whereas the exact McNemar test evaluates the binomial probability under $H_0: p=0.5$ on the 29 discordant pairs $(20, 9)$. At $\mathcal{B}=0.10$, accuracy drops to $78.8\%$ ($197/250$), establishing a significant paired deficit of $\Delta = -12.4$ pp relative to diffuse noise (discordant pairs: 39 vs.\ 8, exact McNemar $p = 5.5 \times 10^{-6}$, $p_{\text{holm}} = 1.1 \times 10^{-5}$, 95\% CI: $[-17.6\text{ pp}, -7.2\text{ pp}]$). Answer completion remains high ($96.4\%$ under Burst vs.\ $98.8\%$ under i.i.d.), and accuracy conditional on emitting an answer line drops from $91.9\%$ to $80.5\%$ ($\Delta = -11.4$ pp), confirming that the deficit predominantly represents errors within derivations that still reach a formatted answer line rather than format omissions.

\paragraph{LLaMA-3.1-8B-Instruct Derivations}
Starting from a clean baseline of $87.2\%$ ($218/250$), diffuse i.i.d.\ noise yields $89.2\%$ at $\mathcal{B}=0.05$ and $84.0\%$ at $\mathcal{B}=0.10$. Subjecting LLaMA to clustered bursts produces significant impairments: at $\mathcal{B}=0.05$, accuracy drops to $81.2\%$ ($\Delta = -8.0$ pp, McNemar $p = 0.0017$, $p_{\text{holm}} = 0.0033$). At $\mathcal{B}=0.10$, accuracy falls to $76.8\%$ ($\Delta = -7.2$ pp, McNemar $p = 0.0198$, $p_{\text{holm}} = 0.0198$, 95\% CI: $[-12.8\text{ pp}, -1.6\text{ pp}]$).

In LLaMA at $\mathcal{B}=0.10$, the answer completion rate drops from $94.0\%$ (i.i.d.) to $90.4\%$ (Burst). Because the scorer falls back to the last numeric value when an explicit \texttt{Answer:} line is absent, 11 items under Burst were scored correct without a final answer line. Among the 36 items where LLaMA succeeded under i.i.d.\ but failed under Burst, 8 items ($22.2\%$) lacked an explicit answer line. When evaluating accuracy strictly conditional on emitting an answer line, the performance gap narrows from $-7.2$ pp (raw) to $-5.0$ pp ($80.1\%$ vs.\ $85.1\%$, Supplementary Table S3).

\paragraph{Realized Head-Token Budgets and Conditional Dose Asymmetry}
Because derivations terminate upon generating EOS tokens, derivations complete in an inferred mean of $\sim 221$--$273$ tokens (estimated from the ratio of realized to scheduled budgets over $T_{\text{max}} = 512$), realizing $\approx 43\%$--$53\%$ of scheduled budgets. In aggregate across all items, mean realized budgets between Burst and i.i.d.\ match closely: on Qwen ($\mathcal{B}=0.10$), $8,897.3$ vs.\ $8,631.3$, while on LLaMA ($\mathcal{B}=0.10$), $6,626.4$ vs.\ $6,437.2$.

However, a conditional dose asymmetry is observed: at $\mathcal{B}=0.10$ on items where Burst failed, derivations ran longer, realizing $+511.1$ (Qwen) and $+826.6$ (LLaMA) more head-tokens than the paired i.i.d.\ arm. On items where Burst succeeded at $\mathcal{B}=0.10$, realized dose differences were markedly smaller ($+200.1$ on Qwen, $-3.4$ on LLaMA). At lower dose $\mathcal{B}=0.05$ on Qwen, failed items averaged $-681.0$ head-tokens relative to i.i.d., demonstrating that post-treatment length variation depends on disruption severity and that budget equality holds strictly at the aggregate population level.

\subsection*{Exploratory Factorial Dissociation}
\label{sec:dissociation_main}

\begin{figure}[htbp]
\centering
\includegraphics[width=0.98\textwidth]{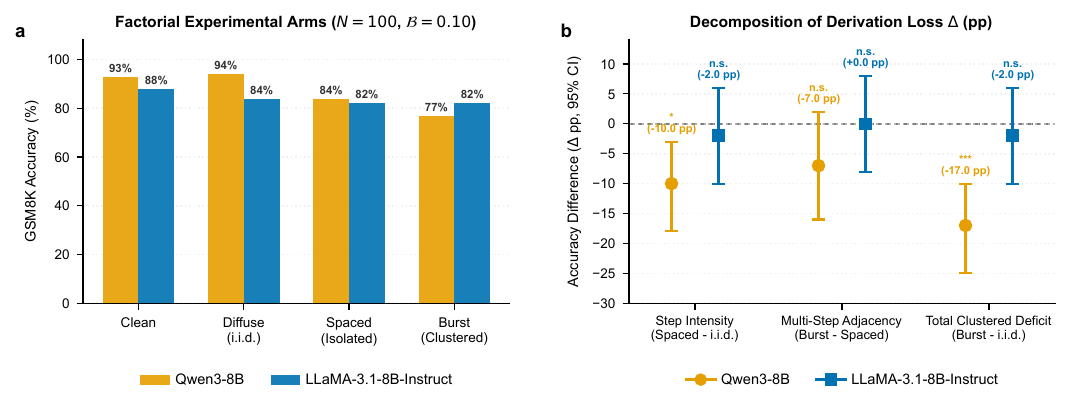}
\caption{\textbf{Factorial dissociation of instantaneous step intensity versus multi-step temporal continuity ($N=100$ paired derivations, nominal onset parameter $\mathcal{B}=0.10$).} \textbf{a}, Factorial arm accuracy on GSM8K across Clean baseline ($M\equiv 1$), Diffuse i.i.d.\ noise, Spaced gapped dropouts (interleaved isolated steps with $\alpha = 0.40$), and Burst consecutive dropouts (consecutive steps $t, t+1, t+2$ with $\alpha = 0.40$). \textbf{b}, Decomposition of derivation loss $\Delta$ (percentage points) with 95\% bootstrap confidence intervals. Instantaneous step intensity ($\text{Spaced} - \text{i.i.d.}$) shows a drop of $-10.0$ pp on Qwen3-8B in this schedule draw ($p_{\text{holm}}=0.0425$). Multi-step adjacency ($\text{Burst} - \text{Spaced}$) accounts for an additional $-7.0$ pp on Qwen, but is statistically inconclusive at $N=100$ ($p_{\text{holm}}=0.1892$, CI includes zero). Total clustered deficit ($\text{Burst} - \text{i.i.d.}$) confirms loss in this draw ($-17.0$ pp, $p_{\text{holm}}=4.6 \times 10^{-5}$). Asterisks: $^{***}\, p_{\text{holm}} < 0.001$, $^{*}\, p_{\text{holm}} < 0.05$, n.s.: not significant.}
\label{fig:dissociation}
\end{figure}

To explore instantaneous per-step density and multi-token temporal continuity, we analyzed the four-arm factorial dissociation experiment at $N=100$ paired items ($\mathcal{B}=0.10$) comparing Clean, Diffuse i.i.d., Spaced (isolated steps with $\alpha = 0.40$), and Burst (consecutive steps with $\alpha = 0.40$) (Fig.~\ref{fig:dissociation}, Table~\ref{tab:dissociation}).

\begin{table}[htbp]
\centering
\footnotesize
\caption{\textbf{Factorial Dissociation Experiment ($N=100$ paired derivations, $\mathcal{B}=0.10$).} Accuracy, realized head budgets, discordant pairs $(b, c)$ where $b=\text{Arm}_2^+ \text{Arm}_1^-$ and $c=\text{Arm}_1^+ \text{Arm}_2^-$ for the paired contrast $(\text{Arm}_1 - \text{Arm}_2)$, exact McNemar $p$-values, and Holm-corrected $p_{\text{holm}}$ across planned contrasts.}
\label{tab:dissociation}
\vspace{2mm}
\resizebox{\textwidth}{!}{%
\begin{tabular}{llccccccc}
\toprule
Model & Arm & Accuracy & Realized Heads & Contrast & $\Delta$ (pp) & Discordant $(b, c)$ & $p_{\text{McNemar}}$ & $p_{\text{holm}}$ \\
\midrule
Qwen3-8B & Clean & 93.0\% & 0.0 & Baseline & --- & --- & --- & --- \\
 & i.i.d. & 94.0\% & 7,485.9 & Baseline i.i.d. & --- & --- & --- & --- \\
 & Spaced & 84.0\% & 7,585.4 & Spaced vs. i.i.d. & $-10.0$ pp & $(13, 3)$ & 0.0213 & 0.0425 \\
 & Burst & 77.0\% & 7,700.4 & Burst vs. Spaced & $-7.0$ pp & $(14, 7)$ & 0.1892 & 0.1892 \\
 & & & & Burst vs. i.i.d. & $-17.0$ pp & $(17, 0)$ & $1.5 \times 10^{-5}$ & $4.6 \times 10^{-5}$ \\
\midrule
LLaMA-3.1-8B-Instruct & Clean & 88.0\% & 0.0 & Baseline & --- & --- & --- & --- \\
 & i.i.d. & 84.0\% & 5,722.1 & Baseline i.i.d. & --- & --- & --- & --- \\
 & Spaced & 82.0\% & 5,719.5 & Spaced vs. i.i.d. & $-2.0$ pp & $(10, 8)$ & 0.8145 & 1.0000 \\
 & Burst & 82.0\% & 5,811.8 & Burst vs. Spaced & $0.0$ pp & $(8, 8)$ & 1.0000 & 1.0000 \\
 & & & & Burst vs. i.i.d. & $-2.0$ pp & $(10, 8)$ & 0.8145 & 1.0000 \\
\bottomrule
\end{tabular}%
}
\end{table}

\paragraph{Exploratory Contrast in Qwen3-8B}
On Qwen3-8B, compressing omissions into isolated single tokens with $\alpha = 0.40$ density (Spaced arm) reduces accuracy from $94.0\%$ (i.i.d.) to $84.0\%$ ($84/100$), generating an acute drop of $-10.0$ pp (discordant pairs: 13 vs.\ 3, exact McNemar $p = 0.0213$, $p_{\text{holm}} = 0.0425$, 95\% CI: $[-18.0\text{ pp}, -3.0\text{ pp}]$). Chaining masks across consecutive tokens ($t, t+1, t+2$ in Burst) further reduces accuracy from $84.0\%$ to $77.0\%$ ($\Delta(\text{Burst} - \text{Spaced}) = -7.0$ pp). However, this contrast does not reach statistical significance after Holm-Bonferroni correction (discordant pairs: 14 vs.\ 7, McNemar $p = 0.1892$, $p_{\text{holm}} = 0.1892$, 95\% CI: $[-16.0\text{ pp}, +2.0\text{ pp}]$). Because the confidence interval spans zero, multi-token temporal continuity cannot be claimed as an established independent factor at $N=100$.

\paragraph{Schedule Differences and Inconclusive Findings on LLaMA-3.1-8B-Instruct}
We explicitly disclose that the dissociation experiment differed from the confirmatory runs in two important respects. First, the generator used a 10-step jump instead of a 7-step jump, scheduling $\sim 13\%$ lower total head budget on the identical 100 items. Second, because of this algorithmic change, random mask placement diverged from the confirmatory run: on these 100 items, Qwen i.i.d.\ scored $94\%$ (compared to $86\%$ in the confirmatory run on the same items), and $25\%$ of items flipped correctness under Burst between runs.

Crucially, on these identical 100 items under the confirmatory schedule, the reference contrast was $\text{Burst} - \text{i.i.d.} = -6.0$ pp on Qwen3-8B (discordant pairs: 12 vs.\ 6, exact McNemar $p = 0.2379$) and $-5.0$ pp on LLaMA-3.1-8B-Instruct (discordant pairs: 13 vs.\ 8, exact McNemar $p = 0.3833$). In contrast, the specific schedule draw in the dissociation experiment yielded $-17.0$ pp on Qwen and $-2.0$ pp on LLaMA. Because each item was tested with a single random schedule realization, standard bootstrapping treats the schedule assignment as fixed, omitting schedule-draw variance. Consequently, the observed single-step density effect ($-10.0$ pp, $p_{\text{holm}} = 0.0425$) must be interpreted strictly as an exploratory, draw-specific finding.

Furthermore, in LLaMA-3.1-8B-Instruct, the reference contrast ($\text{Burst} - \text{i.i.d.}$) failed to reach statistical significance in this $N=100$ draw ($-2.0$ pp, McNemar $p = 0.8145$, $p_{\text{holm}} = 1.000$). Because the reference contrast did not replicate in this specific draw, the equivalence between Spaced and Burst ($82.0\%$ vs.\ $82.0\%$) cannot be interpreted as mechanistic invariance. Rather, this run was underpowered to detect structural decomposition in LLaMA.

\section*{Discussion}

\subsection*{Residual Stream Buffering and Density Differences}
Our findings show a clear distinction between diffuse and concentrated attention head dropouts. When omissions are dispersed uniformly across generation steps as low-density noise, the models exhibit comparatively modest degradation, with Qwen losing $2.4$ pp and LLaMA losing $3.2$ pp relative to clean baseline accuracy at $\mathcal{B}=0.10$. 

However, when head loss concentrates spatiotemporally, performance declines sharply. Silencing $40\%$ of attention heads in the upper layers simultaneously appears to exceed the compensatory capacity of surviving channels, resulting in invalid derivations or missing answer lines.

\subsection*{The Autoregressive Attractor Trap Hypothesis}
To conceptually interpret how transient bursts may trigger persistent derivation errors, we formulate the Autoregressive Attractor Trap Hypothesis as a multi-stage process. 

Multi-step mathematical reasoning naturally forms a sequential dependency graph in which each downstream deduction conditions upon prior tokens. During an attention burst, representational starvation induces disruption through two distinct channels: first, discrete token commitment, where the decoder emits an erroneous intermediate token or premature conclusion into the context history, and second, continuous representational contamination, where perturbed residual stream vectors $x_t^{(l)}$ generate corrupted Key and Value vectors in all subsequent layers $l' > l$, which are written permanently into the KV cache. 

Because autoregressive decoders append both emitted tokens and Key--Value vectors permanently into memory, these corrupted representations become immutable conditioning context \cite{dziri2023faith}. Upon recovery of full attention capacity, subsequent unperturbed heads attend to this corrupted context, trapping the forward trajectory in an erroneous attractor basin from which greedy decoding cannot escape. We emphasize that this mechanism is framed strictly as an interpretative model rather than a proven causal theorem, as surgical KV-cache state rollback was not performed.

\subsection*{Architectural Divergence and Derivation Lengths}
Qwen3-8B and LLaMA-3.1-8B-Instruct exhibit distinct vulnerability profiles. Qwen3-8B features 36 layers and generates longer Chain-of-Thought derivations on GSM8K, yielding higher realized head-token volumes ($8,897.3$ vs.\ $6,626.4$ at $\mathcal{B}=0.10$). We hypothesize post hoc that longer derivation trajectories offer a larger temporal target surface for corrupting omissions, potentially explaining Qwen's steeper decline at $\mathcal{B}=0.10$ and its acute vulnerability to single-step concentration ($\Delta t = 1$). Conversely, LLaMA-3.1-8B-Instruct produces more concise derivations, with a notable fraction of its deficit reflecting format and answer omissions.

\subsection*{Computational Neuroscience Analogies: Biological vs.\ Artificial Networks}
These results provide a conceptual parallel to cortical dynamics during sleep deprivation. In awake mammals, sleep deprivation triggers localized cortical off-states in isolated neuronal assemblies \cite{vyazovskiy2011local, nir2017selective}. In humans, these intrusions predict cognitive lapses during tasks that require continuous, sequential attention \cite{andrillon2021predicting}. 

Artificial autoregressive networks exhibit a functionally related vulnerability: they are resilient to diffuse, unstructured noise, but vulnerable to concentrated disruptions that sever sequential reasoning chains. However, this biological analogy is operational rather than physiological. Artificial attention heads perform dot-product associative routing across token vectors, distinct from electrochemical synaptic transmission and spiking dynamics in biological cortical columns.

\section*{Methods}

\subsection*{Multi-Head Attention and Grouped-Query Formulation}
Let $x_t^{(l)} \in \mathbb{R}^d$ denote the residual stream vector at layer $l \in \{0, \dots, L-1\}$ and generation step $t \in \{1, \dots, T_{\text{max}}\}$, with pre-layer normalization $\bar{x}_t^{(l)} = \text{RMSNorm}(x_t^{(l)}) = \frac{x_t^{(l)}}{\sqrt{\frac{1}{d}\sum_{i=1}^d (x_{t,i}^{(l)})^2 + \epsilon}} \odot \gamma^{(l)}$ \cite{zhang2019root}, where $\gamma^{(l)} \in \mathbb{R}^d$ is a learnable gain parameter and $\epsilon = 10^{-6}$ prevents numerical division by zero. Both models implement Grouped-Query Attention \cite{ainslie2023gqa} with $H=32$ query heads sharing $H_{KV}=8$ Key--Value head pairs ($g(h) = \lfloor h/4 \rfloor$). Attention dot products are computed across prompt tokens and prior decode steps $\tau \in \{1, \dots, P+t\}$ as $(q_t^{(l,h)})^\top k_\tau^{(l,g(h))}$, followed by causal masking and softmax scaling. Layer updates follow standard pre-norm transformer residual connections:
\begin{align}
\tilde{x}_t^{(l)} &= x_t^{(l)} + \widetilde{\text{MHA}}^{(l)}(x_{\le t}) = x_t^{(l)} + \sum_{h=0}^{H-1} M_{l,h,t} W_O^{(l,h)} \text{Attn}^{(l,h)}(\bar{x}_{\le t}), \\
x_t^{(l+1)} &= \tilde{x}_t^{(l)} + \text{FFN}^{(l)}(\text{RMSNorm}(\tilde{x}_t^{(l)})),
\end{align}
where $\widetilde{\text{MHA}}^{(l)}$ denotes the masked multi-head attention operator and $\text{FFN}^{(l)}$ denotes the feed-forward network. Rotary Position Embeddings (RoPE) \cite{su2024roformer} and QK-Norm are omitted from notation for clarity.

\subsection*{Target Layer Subspace and Attention Mask Operator}
Evaluations targeted the upper half of layers in Qwen3-8B ($L=36, H=32, L_{\text{vuln}}=18, N_{\text{vuln}}=576$, layers $\mathcal{L}_{\text{vuln}} = \{18, \dots, 35\}$) and LLaMA-3.1-8B-Instruct ($L=32, H=32, L_{\text{vuln}}=16, N_{\text{vuln}}=512$, layers $\mathcal{L}_{\text{vuln}} = \{16, \dots, 31\}$). Head dropouts were implemented via forward hooks on output projection weights $W_O^{(l,h)}$ during autoregressive generation ($M_{l,h,t} \equiv 1$ for $l \notin \mathcal{L}_{\text{vuln}}$). Silencing head outputs at layer $l \ge l_{\min}$ perturbs the residual stream $x_t^{(l+1)}$. Consequently, Key and Value projections at subsequent layers $l' > l$ ($k_t^{(l', j)}, v_t^{(l', j)}$ for all $j \in \{0, \dots, H_{KV}-1\}$) are computed from perturbed representations and stored in the KV cache. Crucially, while past cache entries ($\tau < t$) are never retroactively overwritten, any corrupted KV entry written during a burst step persists irrevocably in cache memory and continues to influence attention lookups at all subsequent decode steps $\tau' > t$.

\subsection*{Intervention Schedules and Quintile Matching}
The nominal onset parameter $\mathcal{B}$ governs the per-step burst-onset probability at non-refractory steps:
\begin{equation*}
P(\text{start} \mid \text{not refractory}) = \min\left(1.0, \frac{2\mathcal{B}}{\Delta t}\right).
\end{equation*}
The generation horizon was set to $T_{\text{max}} = 512$ tokens. Diffuse i.i.d.\ controls were generated via stratified quintile matching across 5 temporal bins, guaranteeing matched marginal temporal distributions of dropout events between conditions across planned strata.

\subsection*{Inference and Prompt Specification}
Reasoning was evaluated on GSM8K \cite{cobbe2021gsm8k} using a standardized zero-shot user prompt template without a separate system prompt or few-shot exemplars under deterministic greedy decoding ($\text{temperature} = 0.0$, \texttt{do\_sample = False}, \texttt{max\_new\_tokens = 512}). For Qwen3-8B, \texttt{enable\_thinking=False} was configured. Executions ran on Windows 11 with an NVIDIA GeForce RTX 4090 GPU under PyTorch 2.8.0 \cite{paszke2019pytorch} and Hugging Face Transformers (version $\ge 4.51$, unrecorded in run manifests) \cite{wolf2020transformers}. Run manifests recorded \texttt{git\_dirty: true}.

\subsection*{Statistical Analysis}
Significance was tested using exact McNemar tests on paired discordant outcomes $(b, c)$ \cite{mcnemar1947note} with step-down Holm-Bonferroni correction \cite{holm1979simple}, two-sided permutation tests \cite{good2013permutation}, and non-parametric bootstrap resampling (10,000 resamples for 95\% CIs) \cite{efron1994introduction}.

\subsection*{Protocol Registration and Disclosures}
This study was executed following a dated protocol registered before data collection. We disclose all adaptations and deviations: (1) an initial revision of the pre-test specification had already replaced the original candidate models with Qwen3-8B, and the shift from base to instruction-tuned checkpoints specifically applied to LLaMA (\texttt{Llama-3.1-8B-Instruct}) to support zero-shot multi-step reasoning without few-shot prompting, (2) the primary endpoint shifted from per-step recovery latency to end-to-end derivation accuracy, with pairing via 5-quintile stratified marginal position matching, (3) primary sample size allocation of $N=250$ focused on core doses $\mathcal{B} \in \{0.05, 0.10\}$ and $N=60$ on exploratory ablations, (4) reporting exact two-sided McNemar binomial tests on discordant pairs, (5) pre-registered threshold was $\alpha = 0.01$ (LLaMA achieved $p = 0.0198$), (6) LLaMA under Burst showed $9.6\%$ answer format omissions against a registered $2\%$ threshold (a threshold also exceeded in the i.i.d.\ control arms, which exhibited $4.0\%$ at $\mathcal{B}=0.05$ and $6.0\%$ at $\mathcal{B}=0.10$, as well as $7.6\%$ under Burst at $\mathcal{B}=0.05$), (7) replacement of cache rollback surgery with the factorial Spaced dissociation design, and (8) non-monotonic response on Qwen3-8B at $\mathcal{B}=0.20$ as analyzed in Section~\ref{sec:limitations} and Supplementary Note 1.

\section*{Limitations}
\label{sec:limitations}

We discuss several methodological considerations and constraints of our experimental design.

First, regarding budget realization and schedule stochasticity, scheduled budgets were planned across the nominal horizon of $T_{\text{max}} = 512$ tokens. Because GSM8K derivations terminate early upon emitting EOS tokens (averaging an estimated $\sim 221$--$273$ tokens based on realized-to-scheduled ratios), realized budgets represented $43\%$--$53\%$ of scheduled budgets. Furthermore, generating mask schedules via stochastic Bernoulli trials induces item-level variance. The dissociation experiment utilized a 10-step refractory jump rather than the 7-step jump of confirmatory runs, altering scheduled dose by $\sim 13\%$. On the identical 100 items, the confirmatory schedule yielded a reference contrast of $-6.0$ pp on Qwen and $-5.0$ pp on LLaMA, whereas the dissociation draw yielded $-17.0$ pp and $-2.0$ pp, indicating that single-draw evaluations carry substantial schedule variance.

Second, regarding error taxonomy, in LLaMA-3.1-8B-Instruct, $8$ of the $36$ discordant losses ($22.2\%$) at $\mathcal{B}=0.10$ lacked an explicit answer format line, indicating that part of the deficit reflects formatting collapse rather than derivation error. In the duration ablation sweep, the refractory step increment was coupled to $\Delta t + g_{\min}$, so mean scheduled budget scaled with duration. Despite receiving the lowest scheduled dose (13.2k heads), compressing omissions into instantaneous spikes ($\Delta t = 1$) induced a pronounced deficit on Qwen3-8B ($\Delta = -21.7$ pp, $p_{\text{holm}} = 0.0007$), but produced zero observable deficit on LLaMA-3.1-8B-Instruct ($\Delta = 0.0$ pp, $p_{\text{holm}} \ge 0.210$). At elevated budgets ($\mathcal{B} \in \{0.15, 0.20\}$), exploratory sweeps showed non-monotonic responses on Qwen3-8B ($\mathcal{B}=0.15 \to \Delta = -23.3$ pp, $p_{\text{holm}} = 0.0026$ and $\mathcal{B}=0.20 \to \Delta = -15.0$ pp, $p_{\text{holm}} = 0.0225$). At $N=60$, this plateau cannot be statistically distinguished from schedule-draw stochasticity (where Burst rose from 40/60 to 44/60 correct while i.i.d.\ stayed at 54/60 and 53/60). Independently, empirical records show that failed derivations run substantially longer (at $\mathcal{B}=0.20$, fail items under Burst realized $+1,678.1$ additional head-tokens compared to paired i.i.d. derivations), reflecting prolonged derivation trajectories before emitting an EOS token.

Third, regarding mechanistic and architectural scope, in the Spaced control, interleaved tokens were masked using the identical sampled subset of heads across the burst window, while a control resampling fresh heads at each step was not evaluated. In addition, our masking operator intervenes on the output projection $W_O$, zeroing query contextual contributions to the residual stream without performing surgical KV cache state manipulation. While past unmasked tokens retain unperturbed cache vectors, Key and Value projections for current step $t$ in all subsequent layers $l' > l$ ($k_t^{(l', j)}, v_t^{(l', j)}$) are computed from the perturbed residual stream and stored in cache, persisting across all future decoding steps. Finally, reasoning was evaluated zero-shot with thinking disabled on a single mathematical benchmark (GSM8K), and full generation transcripts were not written to disk during evaluation runs.

\section*{Data and Code Availability}
In accordance with double-blind peer review policies, all code, anonymized run manifests, and canonical summary files will be made openly available via an archived public repository upon publication. Plotting scripts reproduce all manuscript figures directly from canonical data.

\bibliographystyle{naturemag}
\bibliography{references}

\begin{thebibliography}{10}
\expandafter\ifx\csname url\endcsname\relax
  \def\url#1{\texttt{#1}}\fi
\expandafter\ifx\csname urlprefix\endcsname\relax\def\urlprefix{URL }\fi
\providecommand{\bibinfo}[2]{#2}
\providecommand{\eprint}[2][]{\url{#2}}

\bibitem{vyazovskiy2011local}
\bibinfo{author}{Vyazovskiy, V.~V.} \emph{et~al.}
\newblock \bibinfo{title}{Local sleep in awake rats}.
\newblock \emph{\bibinfo{journal}{Nature}} \textbf{\bibinfo{volume}{472}},
  \bibinfo{pages}{443--447} (\bibinfo{year}{2011}).

\bibitem{nir2017selective}
\bibinfo{author}{Nir, Y.} \emph{et~al.}
\newblock \bibinfo{title}{Selective neuronal lapses precede human cognitive
  lapses following sleep deprivation}.
\newblock \emph{\bibinfo{journal}{Nature Medicine}}
  \textbf{\bibinfo{volume}{23}}, \bibinfo{pages}{1474--1480}
  (\bibinfo{year}{2017}).

\bibitem{tononi2014sleep}
\bibinfo{author}{Tononi, G.} \& \bibinfo{author}{Cirelli, C.}
\newblock \bibinfo{title}{Sleep and the price of plasticity: from synaptic and
  cellular homeostasis to memory consolidation and integration}.
\newblock \emph{\bibinfo{journal}{Neuron}} \textbf{\bibinfo{volume}{81}},
  \bibinfo{pages}{12--34} (\bibinfo{year}{2014}).

\bibitem{krueger2008sleep}
\bibinfo{author}{Krueger, J.~M.} \emph{et~al.}
\newblock \bibinfo{title}{Sleep as a fundamental property of neuronal
  assemblies}.
\newblock \emph{\bibinfo{journal}{Nature Reviews Neuroscience}}
  \textbf{\bibinfo{volume}{9}}, \bibinfo{pages}{910--919}
  (\bibinfo{year}{2008}).

\bibitem{andrillon2021predicting}
\bibinfo{author}{Andrillon, T.}, \bibinfo{author}{Burns, A.},
  \bibinfo{author}{Mackay, T.}, \bibinfo{author}{Windt, J.} \&
  \bibinfo{author}{Tsuchiya, N.}
\newblock \bibinfo{title}{Predicting lapses of attention with sleep-like slow
  waves through the prism of local sleep}.
\newblock \emph{\bibinfo{journal}{Nature Communications}}
  \textbf{\bibinfo{volume}{12}}, \bibinfo{pages}{2497} (\bibinfo{year}{2021}).

\bibitem{vaswani2017attention}
\bibinfo{author}{Vaswani, A.} \emph{et~al.}
\newblock \bibinfo{title}{Attention is all you need}.
\newblock In \emph{\bibinfo{booktitle}{Advances in Neural Information
  Processing Systems}}, vol.~\bibinfo{volume}{30}, \bibinfo{pages}{5998--6008}
  (\bibinfo{year}{2017}).

\bibitem{wei2022chain}
\bibinfo{author}{Wei, J.} \emph{et~al.}
\newblock \bibinfo{title}{Chain-of-thought prompting elicits reasoning in large
  language models}.
\newblock In \emph{\bibinfo{booktitle}{Advances in Neural Information
  Processing Systems}}, vol.~\bibinfo{volume}{35},
  \bibinfo{pages}{24824--24837} (\bibinfo{year}{2022}).

\bibitem{kojima2022large}
\bibinfo{author}{Kojima, T.}, \bibinfo{author}{Gu, S.~S.},
  \bibinfo{author}{Reid, M.}, \bibinfo{author}{Matsuo, Y.} \&
  \bibinfo{author}{Iwasawa, Y.}
\newblock \bibinfo{title}{Large language models are zero-shot reasoners}.
\newblock In \emph{\bibinfo{booktitle}{Advances in Neural Information
  Processing Systems}}, vol.~\bibinfo{volume}{35},
  \bibinfo{pages}{22199--22213} (\bibinfo{year}{2022}).

\bibitem{cobbe2021gsm8k}
\bibinfo{author}{Cobbe, K.} \emph{et~al.}
\newblock \bibinfo{title}{Training verifiers to solve math word problems}.
\newblock \emph{\bibinfo{journal}{arXiv preprint arXiv:2110.14168}}
  (\bibinfo{year}{2021}).

\bibitem{dziri2023faith}
\bibinfo{author}{Dziri, N.} \emph{et~al.}
\newblock \bibinfo{title}{Faith and fate: Limits of transformers on
  compositionality}.
\newblock In \emph{\bibinfo{booktitle}{Advances in Neural Information
  Processing Systems}}, vol.~\bibinfo{volume}{36},
  \bibinfo{pages}{70287--70311} (\bibinfo{year}{2023}).

\bibitem{wang2023selfconsistency}
\bibinfo{author}{Wang, X.} \emph{et~al.}
\newblock \bibinfo{title}{Self-consistency improves chain of thought reasoning
  in language models}.
\newblock In \emph{\bibinfo{booktitle}{International Conference on Learning
  Representations}} (\bibinfo{year}{2023}).

\bibitem{olsson2022context}
\bibinfo{author}{Olsson, C.} \emph{et~al.}
\newblock \bibinfo{title}{In-context learning and induction heads}.
\newblock \emph{\bibinfo{journal}{Transformer Circuits Thread}}
  (\bibinfo{year}{2022}).

\bibitem{elhage2021mathematical}
\bibinfo{author}{Elhage, N.} \emph{et~al.}
\newblock \bibinfo{title}{A mathematical framework for transformer circuits}.
\newblock \emph{\bibinfo{journal}{Transformer Circuits Thread}}
  (\bibinfo{year}{2021}).

\bibitem{michel2019sixteen}
\bibinfo{author}{Michel, P.}, \bibinfo{author}{Levy, O.} \&
  \bibinfo{author}{Neubig, G.}
\newblock \bibinfo{title}{Are sixteen heads really better than one?}
\newblock In \emph{\bibinfo{booktitle}{Advances in Neural Information
  Processing Systems}}, vol.~\bibinfo{volume}{32},
  \bibinfo{pages}{14014--14024} (\bibinfo{year}{2019}).

\bibitem{voita2019analyzing}
\bibinfo{author}{Voita, E.}, \bibinfo{author}{Talbot, D.},
  \bibinfo{author}{Moiseev, F.}, \bibinfo{author}{Sennrich, R.} \&
  \bibinfo{author}{Titov, I.}
\newblock \bibinfo{title}{Analyzing multi-head self-attention: Specialized
  heads to do the heavy lifting, the rest can be pruned}.
\newblock In \emph{\bibinfo{booktitle}{Proceedings of the 57th Annual Meeting
  of the Association for Computational Linguistics}},
  \bibinfo{pages}{5797--5808} (\bibinfo{year}{2019}).

\bibitem{srivastava2014dropout}
\bibinfo{author}{Srivastava, N.}, \bibinfo{author}{Hinton, G.},
  \bibinfo{author}{Krizhevsky, A.}, \bibinfo{author}{Sutskever, I.} \&
  \bibinfo{author}{Salakhutdinov, R.}
\newblock \bibinfo{title}{Dropout: a simple way to prevent neural networks from
  overfitting}.
\newblock \emph{\bibinfo{journal}{The Journal of Machine Learning Research}}
  \textbf{\bibinfo{volume}{15}}, \bibinfo{pages}{1929--1958}
  (\bibinfo{year}{2014}).

\bibitem{qwen2025qwen3}
\bibinfo{author}{{Qwen Team}}.
\newblock \bibinfo{title}{Qwen3 technical report}.
\newblock \emph{\bibinfo{journal}{arXiv preprint arXiv:2505.09388}}
  (\bibinfo{year}{2025}).

\bibitem{dubey2024llama}
\bibinfo{author}{Dubey, A.} \emph{et~al.}
\newblock \bibinfo{title}{The llama 3 herd of models}.
\newblock \emph{\bibinfo{journal}{arXiv preprint arXiv:2407.21783}}
  (\bibinfo{year}{2024}).

\bibitem{zhang2019root}
\bibinfo{author}{Zhang, B.} \& \bibinfo{author}{Sennrich, R.}
\newblock \bibinfo{title}{Root mean square layer normalization}.
\newblock In \emph{\bibinfo{booktitle}{Advances in Neural Information
  Processing Systems}}, vol.~\bibinfo{volume}{32},
  \bibinfo{pages}{12360--12371} (\bibinfo{year}{2019}).

\bibitem{ainslie2023gqa}
\bibinfo{author}{Ainslie, J.} \emph{et~al.}
\newblock \bibinfo{title}{{GQA}: Training generalized multi-query transformer
  models from multi-head checkpoints}.
\newblock In \emph{\bibinfo{booktitle}{Proceedings of the 2023 Conference on
  Empirical Methods in Natural Language Processing}},
  \bibinfo{pages}{4895--4901} (\bibinfo{year}{2023}).

\bibitem{su2024roformer}
\bibinfo{author}{Su, J.} \emph{et~al.}
\newblock \bibinfo{title}{Ro{F}ormer: Enhanced {T}ransformer with {R}otary
  {P}osition {E}mbedding}.
\newblock \emph{\bibinfo{journal}{Neurocomputing}}
  \textbf{\bibinfo{volume}{568}}, \bibinfo{pages}{127063}
  (\bibinfo{year}{2024}).

\bibitem{paszke2019pytorch}
\bibinfo{author}{Paszke, A.} \emph{et~al.}
\newblock \bibinfo{title}{Py{T}orch: An imperative style, high-performance deep
  learning library}.
\newblock In \emph{\bibinfo{booktitle}{Advances in Neural Information
  Processing Systems}}, vol.~\bibinfo{volume}{32}, \bibinfo{pages}{8024--8035}
  (\bibinfo{year}{2019}).

\bibitem{wolf2020transformers}
\bibinfo{author}{Wolf, T.} \emph{et~al.}
\newblock \bibinfo{title}{Transformers: State-of-the-art natural language
  processing}.
\newblock In \emph{\bibinfo{booktitle}{Proceedings of the 2020 Conference on
  Empirical Methods in Natural Language Processing: System Demonstrations}},
  \bibinfo{pages}{38--45} (\bibinfo{year}{2020}).

\bibitem{mcnemar1947note}
\bibinfo{author}{McNemar, Q.}
\newblock \bibinfo{title}{Note on the sampling error of the difference between
  correlated proportions or percentages}.
\newblock \emph{\bibinfo{journal}{Psychometrika}}
  \textbf{\bibinfo{volume}{12}}, \bibinfo{pages}{153--157}
  (\bibinfo{year}{1947}).

\bibitem{holm1979simple}
\bibinfo{author}{Holm, S.}
\newblock \bibinfo{title}{A simple sequentially rejective multiple test
  procedure}.
\newblock \emph{\bibinfo{journal}{Scandinavian Journal of Statistics}}
  \textbf{\bibinfo{volume}{6}}, \bibinfo{pages}{65--70} (\bibinfo{year}{1979}).

\bibitem{good2013permutation}
\bibinfo{author}{Good, P.}
\newblock \emph{\bibinfo{title}{Permutation tests: a practical guide to
  resampling methods for testing hypotheses}} (\bibinfo{publisher}{Springer
  Science \& Business Media}, \bibinfo{year}{2013}).

\bibitem{efron1994introduction}
\bibinfo{author}{Efron, B.} \& \bibinfo{author}{Tibshirani, R.~J.}
\newblock \emph{\bibinfo{title}{An introduction to the bootstrap}}
  (\bibinfo{publisher}{CRC press}, \bibinfo{year}{1994}).

\end{thebibliography}

\clearpage
\appendix
\begin{center}
{\Large\textbf{Supplementary Material}}
\end{center}
\vspace{5mm}

\section{Supplementary Note 1: Exploratory High-Dose Sweeps (\texorpdfstring{$N=60$}{N=60})}

To explore whether reasoning degradation scales monotonically at elevated intervention intensities, we conducted exploratory sweeps at elevated nominal onset parameters $\mathcal{B} \in \{0.15, 0.20\}$ across $N=60$ paired items for both Qwen3-8B and LLaMA-3.1-8B-Instruct (Table~\ref{tab:supp_high_doses}).

\begin{table}[htbp]
\centering
\footnotesize
\caption{\textbf{Exploratory High-Dose Sweeps ($N=60$ paired GSM8K derivations).} Performance across elevated nominal onset parameters $\mathcal{B} \in \{0.15, 0.20\}$. Columns report: model, experimental arm, nominal onset parameter $\mathcal{B}$, mean scheduled head budget, mean realized head budget, GSM8K accuracy, count, paired effect size contrast $\Delta(\text{Burst} - \text{i.i.d.})$, discordant pairs $(b, c)$ where $b=\text{i.i.d.}^+ \text{Burst}^-$ and $c=\text{Burst}^+ \text{i.i.d.}^-$, exact McNemar $p_{\text{McNemar}}$, and Holm-Bonferroni corrected $p_{\text{holm}}$.}
\label{tab:supp_high_doses}
\vspace{2mm}
\resizebox{\textwidth}{!}{%
\begin{tabular}{llccccccccc}
\toprule
\textbf{Model} & \textbf{Arm} & \textbf{Budget $\mathcal{B}$} & \textbf{Mean Sched} & \textbf{Mean Realized} & \textbf{Accuracy} & \textbf{Count} & \textbf{Paired $\Delta$} & \textbf{Discordant $(b, c)$} & \textbf{$p_{\text{McNemar}}$} & \textbf{$p_{\text{holm}}$} \\
\midrule
\textbf{Qwen3-8B} & Clean & 0.00 & 0.0 & 0.0 & \textbf{90.0\%} & 54/60 & Baseline & --- & --- & --- \\
 & i.i.d. & 0.15 & 21,666 & 11,353.8 & \textbf{90.0\%} & 54/60 & --- & --- & --- & --- \\
 & Burst & 0.15 & 21,666 & 11,381.2 & \textbf{66.7\%} & 40/60 & \textbf{$-23.3$ pp} & $(16, 2)$ & \textbf{0.0013} & \textbf{0.0026} \\
 & i.i.d. & 0.20 & 25,921 & 13,508.2 & \textbf{88.3\%} & 53/60 & --- & --- & --- & --- \\
 & Burst & 0.20 & 25,921 & 14,463.2 & \textbf{73.3\%} & 44/60 & \textbf{$-15.0$ pp} & $(11, 2)$ & \textbf{0.0225} & \textbf{0.0225} \\
\midrule
\textbf{LLaMA-3.1-8B-Instruct} & Clean & 0.00 & 0.0 & 0.0 & \textbf{90.0\%} & 54/60 & Baseline & --- & --- & --- \\
 & i.i.d. & 0.15 & 19,803 & 8,610.5 & \textbf{86.7\%} & 52/60 & --- & --- & --- & --- \\
 & Burst & 0.15 & 19,803 & 8,801.3 & \textbf{78.3\%} & 47/60 & $-8.3$ pp & $(8, 3)$ & 0.2266 & 0.3591 \\
 & i.i.d. & 0.20 & 23,339 & 9,695.0 & \textbf{85.0\%} & 51/60 & --- & --- & --- & --- \\
 & Burst & 0.20 & 23,339 & 10,584.8 & \textbf{75.0\%} & 45/60 & $-10.0$ pp & $(10, 4)$ & 0.1796 & 0.3591 \\
\bottomrule
\end{tabular}%
}
\end{table}

\paragraph{Non-Monotonicity on Qwen3-8B}
On Qwen3-8B, the paired contrast peaks at $\mathcal{B}=0.15$ with an observed deficit of $\Delta = -23.3$ pp ($p_{\text{holm}} = 0.0026$, accuracy dropping from $90.0\%$ to $66.7\%$). However, at $\mathcal{B}=0.20$, the observed deficit contracts to $\Delta = -15.0$ pp ($p_{\text{holm}} = 0.0225$, accuracy $73.3\%$). Rather than premature EOS truncation, analysis of realized budgets reveals that derivations on items where Burst failed ran substantially longer: at $\mathcal{B}=0.20$, failed items under Burst averaged $18,213.1$ realized head-tokens compared to $16,535.1$ on the paired i.i.d. arm (an excess of $+1,678.1$ head-tokens per failed derivation). This indicates that severe disruption leads to prolonged derivation trajectories before emitting an EOS token, rather than premature token truncation.

\paragraph{Response Profile on LLaMA-3.1-8B-Instruct}
On LLaMA-3.1-8B-Instruct, the paired effect sizes remain modest: $\Delta = -8.3$ pp at $\mathcal{B}=0.15$ and $\Delta = -10.0$ pp at $\mathcal{B}=0.20$. At $N=60$, neither contrast reaches family-wise statistical significance after Holm-Bonferroni correction ($p_{\text{holm}} = 0.3591$). We caution that this lack of significance should not be interpreted as evidence of complete representational invariance, as the statistical power at $N=60$ is insufficient to detect moderate effect sizes.

\newpage

\section{Supplementary Note 2: Duration Ablation (\texorpdfstring{$\Delta t \in \{1, 2, 5\}$}{Delta t in {1, 2, 5}})}

To examine how the temporal extent of individual off-state events affects reasoning, we performed an exploratory ablation across burst durations $\Delta t \in \{1, 2, 5\}$ tokens at nominal onset parameter $\mathcal{B}=0.10$ ($N=60$ paired items, Table~\ref{tab:supp_burst_len}).

\begin{table}[htbp]
\centering
\footnotesize
\caption{\textbf{Burst Duration Ablation Sweep ($N=60$ paired GSM8K derivations, $\mathcal{B}=0.10$).} Evaluating temporal duration of individual episodes $\Delta t \in \{1, 2, 5\}$ tokens. Columns report: model, duration $\Delta t$, mean scheduled heads, mean realized heads for Burst (B) and i.i.d.\ (I), observed accuracy, paired effect size contrast $\Delta(\text{Burst} - \text{i.i.d.})$, discordant pairs $(b, c)$ where $b=\text{i.i.d.}^+ \text{Burst}^-$ and $c=\text{Burst}^+ \text{i.i.d.}^-$, exact McNemar $p_{\text{McNemar}}$, and Holm-Bonferroni corrected $p_{\text{holm}}$ (evaluated per model across the 3 duration contrasts).}
\label{tab:supp_burst_len}
\vspace{2mm}
\resizebox{\textwidth}{!}{%
\begin{tabular}{llccccccccc}
\toprule
\textbf{Model} & \textbf{Duration $\Delta t$} & \textbf{Mean Sched} & \textbf{Realized (B)} & \textbf{Realized (I)} & \textbf{Burst Acc} & \textbf{i.i.d. Acc} & \textbf{Paired $\Delta$} & \textbf{Discordant $(b, c)$} & \textbf{$p_{\text{McNemar}}$} & \textbf{$p_{\text{holm}}$} \\
\midrule
\textbf{Qwen3-8B} & $\Delta t = 1$ & 13,214 & 7,122.3 & 6,699.9 & \textbf{71.7\%} & \textbf{93.3\%} & \textbf{$-21.7$ pp} & $(13, 0)$ & \textbf{0.0002} & \textbf{0.0007} \\
 & $\Delta t = 2$ & 15,502 & 7,969.5 & 7,990.1 & \textbf{78.3\%} & \textbf{90.0\%} & $-11.7$ pp & $(9, 2)$ & 0.0654 & 0.1309 \\
 & $\Delta t = 5$ & 18,323 & 9,644.7 & 9,267.1 & \textbf{75.0\%} & \textbf{85.0\%} & $-10.0$ pp & $(10, 4)$ & 0.1796 & 0.1796 \\
\midrule
\textbf{LLaMA-3.1-8B-Instruct} & $\Delta t = 1$ & 11,709 & 5,217.2 & 5,125.5 & \textbf{85.0\%} & \textbf{85.0\%} & $0.0$ pp & $(4, 4)$ & 1.0000 & 1.0000 \\
 & $\Delta t = 2$ & 14,104 & 6,225.2 & 6,074.3 & \textbf{80.0\%} & \textbf{90.0\%} & $-10.0$ pp & $(7, 1)$ & 0.0703 & 0.2109 \\
 & $\Delta t = 5$ & 15,290 & 6,662.5 & 6,434.1 & \textbf{88.3\%} & \textbf{95.0\%} & $-6.7$ pp & $(6, 2)$ & 0.2891 & 0.5781 \\
\bottomrule
\end{tabular}%
}
\end{table}

\paragraph{Single-Step Concentration vs.\ Duration Confounding}
In Qwen3-8B, compressing omissions into single-step instantaneous spikes ($\Delta t = 1$) produced the sharpest observed contrast: accuracy dropped to $71.7\%$ under Burst versus $93.3\%$ under i.i.d.\ ($\Delta = -21.7$ pp, exact McNemar $p = 0.0002$, $p_{\text{holm}} = 0.0007$). In LLaMA-3.1-8B-Instruct, the $\Delta t = 1$ contrast was exactly zero ($85.0\%$ vs.\ $85.0\%$, $\Delta = 0.0$ pp, $p_{\text{holm}} = 1.0000$), while contrasts at $\Delta t \in \{2, 5\}$ did not achieve statistical significance after multiple testing correction ($p_{\text{holm}} \ge 0.210$).

Importantly, we disclose a methodological limitation of this ablation. In this run, the schedule generator advanced by $\Delta t + g_{\min}$ steps after each burst (with $g_{\min} = 4$). Because $\Delta t$ directly modified the refractory step increment, the empirical mean scheduled heads rose across duration conditions (in Qwen: 13.2k at $\Delta t=1$, 15.5k at $\Delta t=2$, 18.3k at $\Delta t=5$). Consequently, duration was partially confounded with total scheduled dose in this run. In addition, post-treatment realized length differences meant that under $\Delta t=1$ on Qwen, mean realized Burst dose slightly exceeded i.i.d.\ by $\sim 6\%$ ($7,122$ vs.\ $6,700$ heads), and under $\mathcal{B}=0.20$ by $7\%$--$9\%$, consistent with prolonged derivation paths upon disruption.

\newpage

\section{Supplementary Note 3: Quantitative Analysis of Derivation Completion and Error Modes}

\paragraph{Data Logging Disclosure}
In the experimental pipeline, full token-by-token generation transcripts were not written to disk. Instead, structured metadata records capturing item-level derivation characteristics were logged directly to \texttt{results.jsonl}. To evaluate the nature of reasoning failures without relying on reconstructed or synthetic transcripts, Table~\ref{tab:supp_error_modes} presents an audited breakdown of final-answer presence (\texttt{has\_final\_answer}), raw accuracy, and conditional accuracy defined as:
\begin{equation}
\text{Acc}_{\text{cond}} = P(\text{correct} \mid \text{has\_final\_answer}) = \frac{\sum (\text{correct} \land \text{has\_final\_answer})}{\sum \text{has\_final\_answer}}.
\end{equation}

\begin{table}[htbp]
\centering
\footnotesize
\caption{\textbf{Derivation Completion and Answer Presence Across Experimental Arms.} Audited metrics computed directly from recorded \texttt{results.jsonl} files. Columns report: model, evaluation suite, experimental arm, nominal onset parameter $\mathcal{B}$, sample size $N$, final answer emission rate (\texttt{has\_final\_answer}), raw task accuracy, and accuracy conditional on emitting a final answer line.}
\label{tab:supp_error_modes}
\vspace{2mm}
\resizebox{\textwidth}{!}{%
\begin{tabular}{llccccccc}
\toprule
\textbf{Model} & \textbf{Experiment} & \textbf{Arm} & \textbf{Budget $\mathcal{B}$} & \textbf{$N$} & \textbf{Final Answer Rate} & \textbf{Raw Acc} & \textbf{Conditional Acc} \\
\midrule
\textbf{Qwen3-8B} & Confirmatory & Clean & 0.00 & 250 & 99.6\% & 93.6\% & 94.0\% \\
 & Confirmatory & i.i.d. & 0.05 & 250 & 98.4\% & 92.4\% & 93.5\% \\
 & Confirmatory & Burst & 0.05 & 250 & 98.8\% & 88.0\% & 88.7\% \\
 & Confirmatory & i.i.d. & 0.10 & 250 & 98.8\% & 91.2\% & 91.9\% \\
 & Confirmatory & Burst & 0.10 & 250 & 96.4\% & 78.8\% & 80.5\% \\
\midrule
 & Dissociation & Clean & 0.00 & 100 & 100.0\% & 93.0\% & 93.0\% \\
 & Dissociation & i.i.d. & 0.10 & 100 & 99.0\% & 94.0\% & 93.9\% \\
 & Dissociation & Spaced & 0.10 & 100 & 100.0\% & 84.0\% & 84.0\% \\
 & Dissociation & Burst & 0.10 & 100 & 96.0\% & 77.0\% & 80.2\% \\
\midrule
\textbf{LLaMA-3.1-8B-Instruct} & Confirmatory & Clean & 0.00 & 250 & 98.8\% & 87.2\% & 87.9\% \\
 & Confirmatory & i.i.d. & 0.05 & 250 & 96.0\% & 89.2\% & 90.0\% \\
 & Confirmatory & Burst & 0.05 & 250 & 92.4\% & 81.2\% & 84.0\% \\
 & Confirmatory & i.i.d. & 0.10 & 250 & 94.0\% & 84.0\% & 85.1\% \\
 & Confirmatory & Burst & 0.10 & 250 & 90.4\% & 76.8\% & 80.1\% \\
\midrule
 & Dissociation & Clean & 0.00 & 100 & 99.0\% & 88.0\% & 87.9\% \\
 & Dissociation & i.i.d. & 0.10 & 100 & 92.0\% & 84.0\% & 85.9\% \\
 & Dissociation & Spaced & 0.10 & 100 & 91.0\% & 82.0\% & 83.5\% \\
 & Dissociation & Burst & 0.10 & 100 & 94.0\% & 82.0\% & 83.0\% \\
\bottomrule
\end{tabular}%
}
\end{table}

\paragraph{Separation of Formatting Derailment from Derivation Discrepancy}
The data in Table~\ref{tab:supp_error_modes} demonstrate two distinct response characteristics across models:
\begin{enumerate}
  \item \textbf{Qwen3-8B Derivations:} In Qwen3-8B at $\mathcal{B}=0.10$, answer emission rates remain high ($96.4\%$ under Burst vs.\ $98.8\%$ under i.i.d.). The conditional accuracy drops from $91.9\%$ to $80.5\%$ ($\Delta = -11.4$ pp), closely mirroring the raw task accuracy decline of $-12.4$ pp. This confirms that in Qwen, reasoning failure predominantly reflects derivation errors within completed derivation chains rather than format omissions.
  \item \textbf{LLaMA-3.1-8B-Instruct Derivations:} In LLaMA-3.1-8B-Instruct at $\mathcal{B}=0.10$, the answer emission rate declines from $94.0\%$ (i.i.d.) to $90.4\%$ (Burst). Because the GSM8K scorer falls back to the last numerical value when an explicit \texttt{Answer:} line is absent, 11 items under Burst were scored correct without a final answer line. Among the 36 items where LLaMA succeeded under i.i.d.\ but failed under Burst, 8 items ($22.2\%$) lacked an explicit answer line. When conditioning on answer presence, the performance drop narrows from $-7.2$ pp (raw) to $-5.0$ pp ($80.1\%$ vs.\ $85.1\%$). A similar pattern occurs at $\mathcal{B}=0.05$, where 8 of the 29 discordant losses ($27.6\%$) likewise lacked an answer format line, indicating consistent susceptibility to formatting derailment in LLaMA.

\end{enumerate}

\newpage

\section{Supplementary Note 4: Experimental Battery \& Prompt Specification}

\subsection{Standardized Zero-Shot Prompt Template}
All mathematical reasoning evaluations utilized the standardized zero-shot user prompt template without a separate system prompt:

\begin{quote}
\small\ttfamily\raggedright
Solve the following grade-school math problem. Work through it step by step, then end with a final line of the form "Answer: <number>", where <number> is just the number (no units).

\medskip
Problem: \{question\}
\end{quote}

The evaluation framework avoids few-shot exemplars, evaluating zero-shot reasoning directly. For Qwen3-8B, decoding was executed with \texttt{enable\_thinking=False}.

\subsection{Response Scoring and Answer Parsing}
Responses are scored numerically using standard GSM8K evaluation rules:
\begin{enumerate}
  \item The response is parsed with a regular expression targeting explicit answer declarations:
  \begin{quote}
  \small\ttfamily\raggedright
  answer(?:\textbackslash s+is\textbackslash b)?[\textbackslash s:=*\_`(\textbackslash [\{\textbackslash \textbackslash \$<]*([-\textbackslash u2212]?\textbackslash d[\textbackslash d,]*(?:\textbackslash .\textbackslash d+)?)
  \end{quote}
  \item If no match is found, the scorer searches for boxed LaTeX expressions:
  \begin{quote}
  \small\ttfamily\raggedright
  \textbackslash \textbackslash boxed\{[\textbackslash s\textbackslash \textbackslash \$]*([-\textbackslash u2212]?\textbackslash d[\textbackslash d,]*(?:\textbackslash .\textbackslash d+)?)\}
  \end{quote}
  \item If neither marker is found, the scorer falls back to the last numerical value in the generated sequence.
  \item The extracted float is compared against the gold reference target extracted from the GSM8K test split. An exact match within absolute tolerance of $10^{-6}$ is scored as $1.0$, and $0.0$ otherwise.
\end{enumerate}

\subsection{Hardware, Software Environment and Versioning}
\begin{itemize}
  \item \textbf{Host System:} Dedicated workstation running Windows 11 (build 10.0.26200-SP0).
  \item \textbf{GPU Hardware:} NVIDIA GeForce RTX 4090 (24GB GDDR6X VRAM).
  \item \textbf{Software Stack:} Python 3.12.5, PyTorch 2.8.0+cu129, Hugging Face Transformers (version $\ge 4.51$, unrecorded in run manifests).
  \item \textbf{Precision:} Native \texttt{bfloat16} tensor execution.
  \item \textbf{Decoding Configuration:} Deterministic greedy decoding with parameters: \texttt{temperature = 0.0}, \texttt{do\_sample = False}, and \texttt{max\_new\_tokens = 512}.
  \item \textbf{Seed Invariance:} Evaluation items are paired across experimental conditions with identical problem-level random seeds (\texttt{seed = 42 + item\_idx * 1000}).
  \item \textbf{Manifest Versioning:} In all run manifests, \texttt{git\_dirty: true} was recorded, reflecting active local development of intervention hooks during execution.
\end{itemize}

\end{document}